# When Search Teaches Style: Causal Internalization of Tactical Priors in AlphaZero

Ruitong Li[1,*], Binjie Guo[2,*], Aisheng Mo[2], Guowei Su[2], Han Wang[3], Jie Li[4], and Ru Zhang[2]

[1]The University of Hong Kong
[2]Zhejiang University
[3]Dalian University of Technology
[4]Independent Researcher
[*]These authors contributed equally.

**Abstract**

AlphaZero is normally evaluated as one agent: a policy–value network fused with Monte Carlo tree search. That fusion hides a causal question. When self-play search is given a useful prior, does the network absorb the induced behavior, or does the behavior stay rented from search at test time? We answer with Cross-Phase Prior Intervention (CPI), which switches a root-level tactical prior on and off independently during training and during evaluation, separating the prior's online effect from the learned residual it leaves in the weights. The endpoint is deliberately narrow: how often a network discharges a forced defensive obligation when no search-time guidance is available. On a sealed one-shot final test in 9×9 Gomoku and 19×19 Go, deleting the prior still leaves a large residual—response rises from 13.8% to 26.3% in Gomoku and from 0.6% to 33.8% in Go—and soft reweighting teaches as well as hard action restriction, so pruning legal actions is not the mechanism. The same cross bounds the claim: re-enabling the prior restores nearly 100% response, leaving dependence gaps of 73.7 and 65.8 points. A matched-position evaluation localizes the residual to trained geometry—absent at the shared initialization, emerging over training, worth +17.3 points on in-distribution defenses but only +1.3 on structurally novel ones. Search is therefore best read as a training-time behavioral curriculum whose lessons are real, partial, and geometry-bound, and online competence and internalized competence are different estimands that a diagonal ablation cannot tell apart.

## 1 Introduction

AlphaZero-style systems couple self-play, a policy–value network, and Monte Carlo tree search (MCTS) into a single learning loop [Silver et al., 2016, 2017, 2018]. That loop has proved fertile ground for search improvements: alternative action-selection rules, domain priors, learned dynamics, and cheaper root exploration [Danihelka et al., 2022, Schrittwieser et al., 2020, Hubert et al., 2021, Ye et al., 2021]. Almost all of them are judged the same way, by the strength of the complete agent. When such a modification improves play, the published number cannot say whether the network learned a better policy or whether search simply repaired the same network at inference time.

The distinction is not cosmetic. A learned behavior survives when search is shortened, distilled, or removed; an online correction does not. It also changes what an improved self-play curve means, because search here is not only an inference procedure: it chooses which states enter the replay buffer and which action distributions become targets. Curriculum learning, imitation, dataset aggregation, and policy distillation each describe a piece of this [Bengio et al., 2009, Ross et al., 2011, Hinton et al., 2015, Rusu et al., 2016], but AlphaZero search plays teacher, data generator, and actor at once, and an end-to-end win rate cannot attribute a gain to any one role.

We turn this ambiguity into a measurement. CPI controls a deterministic root-level tactical prior as two separate factors: whether it is present while the agent trains, and whether it is present while the agent is evaluated. Because the two switches are independent, the resulting 3 × 3 cross (Fig. 1) contains cells that a conventional ablation never visits. Evaluating a guided network with the prior deleted measures behavior that must be stored in the weights, since no tactical computation is available at test time. Evaluating an unguided network with the prior enabled measures the opposite quantity: capability supplied entirely online. The diagonal that the literature normally reports is exactly the sum of the two, plus an interaction, and we show below that this decomposition is an identity rather than an approximation.

Our test bed is two games with different geometry and branching: 9×9 Gomoku and 19×19 Go. At the MCTS root, a rule-based tactical module either softly reweights the prior over legal actions or restricts search to a forcing set—completing or blocking a five and creating or blocking an open four in Gomoku, capturing or rescuing a group in atari in Go. Crucially, the module's output is never a network input and never reaches the value head, so any behavior that persists after the module is removed had to be transported through search targets and self-play trajectories. Throughout, style means only the action preferences a network expresses when no search-time guidance is available; the pre-registered endpoint is defensive obligation response, and we make no claim about aesthetic or strategic style more broadly.

Three findings follow. The residual is large and replicates across games: with the prior deleted, force guidance raises held-out defensive response by 12.6 points in Gomoku (13.8% → 26.3%) and 33.2 points in Go (0.6% → 33.8%), and soft reweighting matches it (+12.7 and +31.1), inverting the intuition that hard constraints teach hardest. The residual is also far from complete: re-enabling the forcing prior lifts response to nearly 100%, so six-sevenths of the Gomoku behavior and two-thirds of the Go behavior remain rented from search. And it is structured—absent at the shared initialization, emerging over checkpoints, then collapsing under structural shift, where the force-guided advantage falls from +17.3 points on in-distribution defensive templates to +1.3 on geometries the curriculum never saw.

We contribute a **design** that crosses training-time and evaluation-time interventions so the online and learned components of a search prior are separately identified rather than jointly reported; a **decomposition** of the diagonal ablation into the learned residual, the online effect, and a cross-term — elementary algebra whose force is entirely negative, since no assumption removes the cross-term and the off-diagonal cells therefore cannot be inferred from published numbers but have to be run; **cross–game evidence with explicit bounds**, since a large dependence gap and a collapse under structural shift limit what internalization means here; and a **verifiable protocol**—hash- verified shared initialization, an intervention-independent oracle, a verified zero-call guarantee under Prior-Off, and a one-shot final-test ledger.

## 2 Related Work

MCTS allocates simulations using value estimates and bandit exploration [Coulom, 2006, Kocsis and Szepesvári, 2006, Browne et al., 2012]. AlphaGo, AlphaGo Zero, and AlphaZero train the policy toward search visit counts, making search an implicit policy-improvement operator [Silver et al., 2016, 2017, 2018]; Expert Iteration and regularized-policy formulations make the teacher–student reading explicit [Anthony et al., 2017, Grill et al., 2020]. MuZero and its relatives strengthen the loop through learned dynamics, root action selection, action sampling, or auxiliary objectives [Schrittwieser et al., 2020, Danihelka et al., 2022, Hubert et al., 2021, Ye et al., 2021]. All of them evaluate the search-coupled agent. We ask the complementary question: what is left when the search teacher walks away.

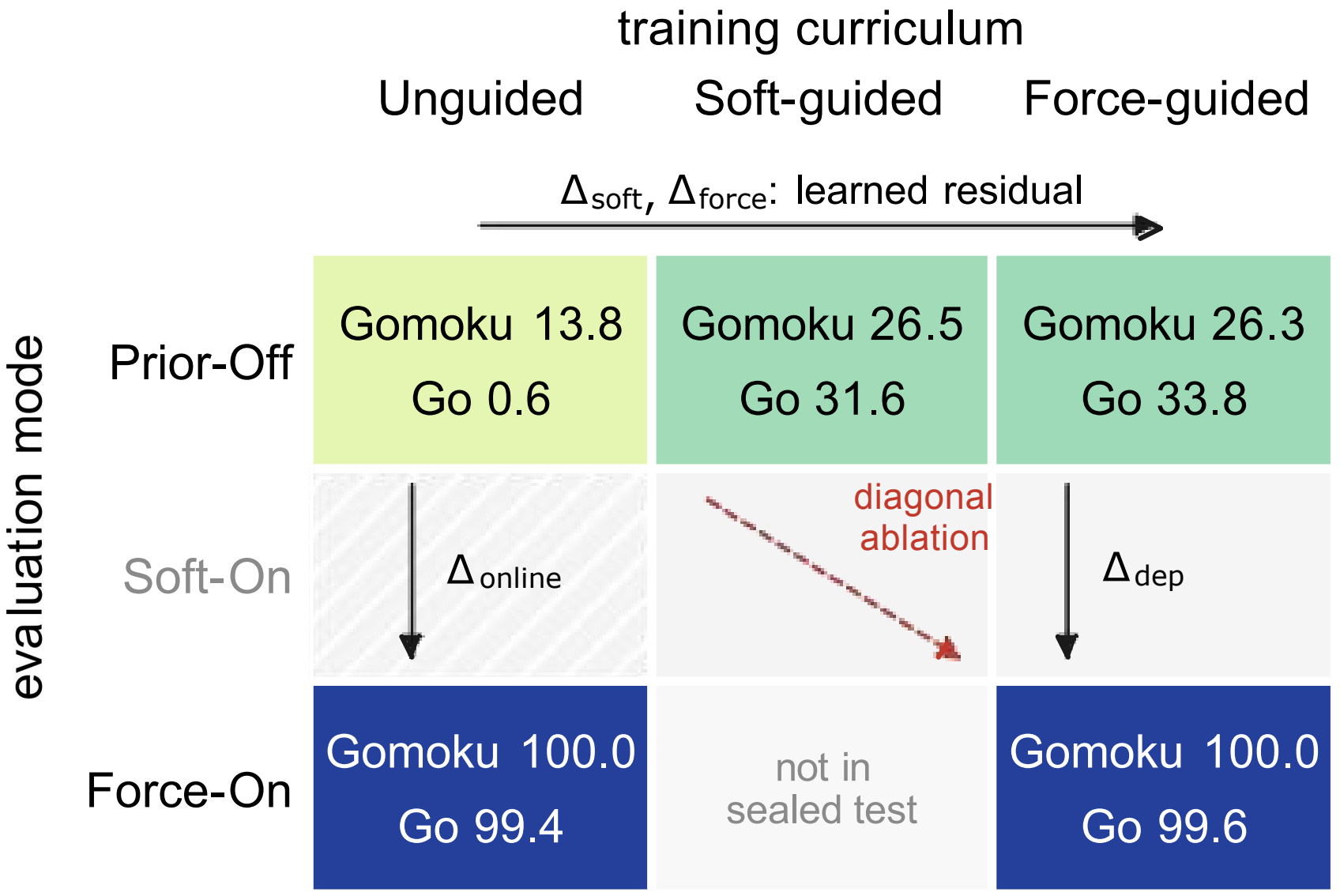

Figure 1: The CPI cross: training curriculum (columns) crossed with evaluation mode (rows). Shaded cells are the five pre-registered final-test cells; entries are defensive obligation response (%) at checkpoint 40 over three seeds ($n$ = 3,831 Gomoku, $n$ = 10,194 Go per cell). The four hatched cells were never opened, and the arrows drawn inside them show why they need not be: every estimand in Eq. 6 is a contrast between two shaded cells—learned residuals along the Prior-Off row, the online effect down the Unguided column, the remaining dependence down the Force-Guided column. A conventional ablation reports only the corner-to-corner diagonal, which confounds all three. Soft-On was used freely in development and validation and is reported in Fig. 2(c); pre-registration kept it out of the confirmatory protocol.

The nearest conceptual relatives concern curricula, imitation, and knowledge transfer. Curriculum and self-paced learning order the training examples [Bengio et al., 2009, Kumar et al., 2010]; distillation transfers a teacher's distribution [Hinton et al., 2015, Rusu et al., 2016]; dataset aggregation collects learner-induced states [Ross et al., 2011]; reward shaping injects domain knowledge into the objective [Ng et al., 1999]. Our intervention is weaker than all of these: it alters neither reward, observation, architecture, nor loss, but only the root prior—and hence only the search targets and the states subsequently visited. Internalization is therefore a property of the closed training loop rather than of an added supervision signal, which is what makes the crossed design necessary.

A question is also only as good as its protocol. Deep-RL conclusions shift with seeds, endpoints, and aggregation choices [Henderson et al., 2018, Agarwal et al., 2021, Patterson et al., 2023], and held-out tests routinely expose failures that training scores hide [Zhang et al., 2018, Cobbe et al., 2019, Kirk et al., 2023, Wu, 2019]. We treat both switches as explicit interventions [Pearl, 2009, Imbens and Rubin, 2015], which is what separates behavior stored in weights from behavior supplied online.

# 3 Cross–Phase Prior Intervention

The base agent is a standard AlphaZero implementation: a residual policy–value network trained on the usual sum of policy cross-entropy, value regression, and weight decay, with PUCT search. Both games use 17 input planes (eight past positions per color plus side to move); Go uses a smaller, independently sized network so that a fully crossed experiment on 19×19 remains affordable. Compute is matched across curricula within each game, and no causal contrast is ever formed across games.

## 3.1 The Guidance Operators

Let $A(s)$ be the legal actions at root state $s$ and let $p_\theta(\cdot \mid s)$ be the network prior normalized on $A(s)$. A deterministic tactical module returns a threat score $r(s, a) \in [0, W]$—the value of the best tactic the action realizes or refutes, read off a fixed tier table, with $W$ the value of an immediate win—and a forcing set $M(s) \subseteq A(s)$ of actions whose omission loses immediately. In Gomoku $M(s) = \{a : r(s, a) \geq \tau\}$ for a fixed tier $\tau$, which collects completing a five, blocking an opponent's five, and creating or blocking an open four; in Go a tier contributes to $M(s)$ only when its candidate is unique, so a genuine choice between captures or between rescues is never pruned. Either way, membership in $M(s)$ requires a strictly positive tier, so $r(s, a) > 0$ for every $a \in M(s)$. From these the module forms a logit bonus

$$b(s,a) = \frac{\lambda}{W}\Big[r(s,a) + \tau\,\mathbf{1}\{a \in M(s)\}\Big], \tag{1}$$

with $\lambda > 0$ a fixed strength, so $b$ is bounded in $[0, \lambda(1 + \tau/W)]$, an action with no tactical content receives $b = 0$ and leaves $p_\theta$ untouched, and the indicator vanishes whenever $M(s) = \varnothing$. Here $\tau$ sizes the membership bonus and, in Gomoku only, doubles as the threshold that defines $M(s)$. All constants are listed in the technical appendix.

The three evaluation modes are three maps from the same network prior to a root prior $q_g(\cdot \mid s;\theta)$ supported on a root action set $A_g(s)$. Prior-Off leaves both untouched: $q_O = p_\theta$ on $A_O(s) = A(s)$. Soft-On tilts the prior but keeps full support, $A_S(s) = A(s)$,

$$q_S(a \mid s;\theta) = \frac{p_\theta(a \mid s)\, e^{b(s,a)}}{\sum_{u \in A(s)} p_\theta(u \mid s)\, e^{b(s,u)}}. \tag{2}$$

Force-On restricts the root to the forcing set when one exists, $A_F(s) = M(s)$, and otherwise falls back to Eq. 2 on $A_F(s) = A(s)$. Writing $R_M(s) = \sum_{u \in M(s)} r(s, u)$,

$$q_F(a \mid s;\theta) = \begin{cases} r(s,a)/R_M(s), & M(s) \neq \emptyset,\ a \in M(s), \\ 0, & M(s) \neq \emptyset,\ a \notin M(s), \\ q_S(a \mid s;\theta), & M(s) = \emptyset. \end{cases} \tag{3}$$

This is a well-defined distribution: $r > 0$ on $M(s)$ gives $R_M(s) > 0$ whenever $M(s) \neq \varnothing$, and the third branch is Eq. 2, normalized on $A(s)$. Two properties matter later. First, Eq. 3 does not merely mask $p_\theta$; on a nonempty forcing set it replaces the network prior by the module's own scores. Near-perfect Force-On accuracy is therefore guaranteed by construction and serves below only as a positive control and as the reference level for a dependence gap. Second, guidance is confined to the root: internal tree nodes always expand with $p_\theta$, and $r$, $b$, and $M$ never enter the feature planes or the value head. During self-play the root prior is additionally mixed with Dirichlet noise after the operator is applied, on the possibly restricted action set. Evaluation is noise-free: writing $N_g(s, a;\theta)$ for the root visit count that PUCT accumulates from $q_g$ under a fixed simulation budget, the agent plays the most-visited root action,

$$\hat{a}_{\theta,g}(s) = \arg\max_{a \in A_g(s)} N_g(s, a;\theta). \tag{4}$$

## 3.2 Factorial Design and Estimands

The three training curricula—Unguided, Soft-Guided, Force-Guided—differ only in which operator drives self-play search. Within each seed they start from a single shared untrained initialization and train for 40 complete iterations; checkpoints are kept after iterations 0, 2, 5, 10, 20, and 40. Every checkpoint is then evaluated under all three modes on identical position sets with a common budget of 32 simulations, so evaluation compute never differs across arms.

Write $c \in \{U, S, F\}$ for the Unguided, Soft-Guided, and Force-Guided curricula and $g \in \{O, S, F\}$ for the Prior-Off, Soft-On, and Force-On modes. Let $\theta_{c,j,k}$ be the checkpoint-k network of curriculum c under seed j, let $\hat{a}_{\theta,g}(s)$ be the action Eq. 4 selects, and let $Y(s) \subseteq A(s)$ be the complete set of correct actions returned by an independent oracle. Over the pooled index D of applicable seed–position pairs, the reported response is

$$R_c^g(k) = \frac{1}{|\mathcal{D}|} \sum_{(j,s)\in\mathcal{D}} \mathbf{1}\left\{\hat{a}_{\theta_{c,j,k},g}(s) \in Y(s)\right\}. \tag{5}$$

Scoring against a set rather than a single move matters: several defenses are often simultaneously valid, and a unique answer key would penalize correct play. Abbreviating $R_c^g \equiv R_c^g(40)$, the pre-declared contrasts are

$$\Delta_{\text{online}} = R_U^F - R_U^O, \qquad \Delta_{\text{soft}} = R_S^O - R_U^O,$$
$$\Delta_{\text{force}} = R_F^O - R_U^O, \qquad \Delta_{\text{dep}} = R_F^F - R_F^O. \tag{6}$$

$\Delta_{\text{online}}$ is capability the module supplies at inference without touching the weights; $\Delta_{\text{soft}}$ and $\Delta_{\text{force}}$ are learned residuals, and because both hold evaluation fixed at O neither can contain an online contribution; $\Delta_{\text{dep}}$ is what training failed to internalize. Because all four are differences among the same four cells, the diagonal comparison that a conventional ablation reports decomposes exactly:

$$\underbrace{R_F^F - R_U^O}_{\text{diagonal ablation}} = \Delta_{\text{force}} + \Delta_{\text{dep}}$$
$$= \Delta_{\text{force}} + \Delta_{\text{online}} + I, \tag{7}$$

where the interaction is $I := \Delta_{\text{dep}} - \Delta_{\text{online}}$. The algebra is elementary—bookkeeping on four cells—and that is exactly what makes it binding: because Eq. 7 holds identically, I cannot be argued away by appealing to approximate additivity. It is nonzero precisely when the online value of the prior differs between an unguided and a guided network, the generic case once training has changed the policy. So one cannot recover a learned residual by subtracting a separately measured inference-time gain from a published end-to-end number; the off-diagonal cells have to be run.

## 3.3 Identification and Negative Controls

Reading $\Delta_{\text{soft}}$ and $\Delta_{\text{force}}$ causally requires three conditions, each paired with a control we can check.

**Pre-training exchangeability.** The arms must be interchangeable before any gradient step. We save one untrained initialization per seed and verify a canonical hash over ordered parameter names, shapes, dtypes, and raw bytes, and no arm is ever warm-started from another. The check is behavioral as well as structural: at checkpoint 0 the three arms agree on matched positions to within 3 of 271 correct responses in-distribution and 3 of 1,500 out-of-distribution (Fig. 2, checkpoint 0).

**No leakage of evaluation mode into the network.** Switching modes must change only the root search. Under Prior-Off the module's call counter is required to be exactly zero, verified over every Prior-Off defense record; and for a fixed checkpoint, moving from Prior-Off to Force-On must leave network policy entropy and network tactical mass bitwise unchanged. Table 1 shows this invariance at final test, which rules out an implementation in which "turning search on" silently swaps features or weights.

**Treatment-independent outcome.** The oracle must not be the intervention in disguise. The tactical labeler lives in a module statically forbidden from importing the intervention code; it enumerates correct actions from the game rules and returns the complete set. Golden tests verify one direction only—every forcing action the intervention proposes must be legal and accepted by the oracle—while allowing the oracle's set to be strictly larger. Without this separation, near-perfect Force-On accuracy would be circular by construction.

Two further patterns would undercut an internalization reading, and neither holds. An initialization artifact would already be present at checkpoint 0, but the residual is absent there and emerges with training (Fig. 2); generic confidence sharpening would make response track policy entropy, but entropy moves non-monotonically with response both within and across games (Table 1).

## 3.4 Data, Splits, and Reporting

The recorded archive contains 701,483 decision records (539,661 Gomoku, 161,822 Go), of which 534,074 carry an applicable tactical label. Position families are canonicalized under game symmetries before splitting, and no family crosses the development, validation, and test partitions. Defensive validation is deliberately two-layered and the layers are never pooled into one generalization number: an in-distribution layer of held-out instances of trained families, and a structural-OOD layer of tactical geometries appearing in no training family—sealed and broken fours, corner obligations, multi-threat states, multi-stone atari chains, counter-captures, and positions with several valid saves. The final test was sealed before the analysis lock and queried exactly once, after code hashes, hypotheses, and manifests were frozen; the ledger records the lock identifier, the five cells opened, and the closeout time. It contains 19,155 Gomoku and 50,970 Go evaluations across three formal seeds.

Two reporting choices deserve emphasis. First, the archive is not fully balanced across evaluation modes for every arm, so an archive-wide average would compare arms over different mode mixtures. We therefore report only matched cells: the pre-registered cells of Table 1, or the fully populated evaluation of Fig. 2, in which all three arms are scored on identical positions under identical modes. Second, the frozen analysis plan specifies hierarchical resampling over seed and template family, with Holm correction across primary contrasts and Benjamini–Hochberg control for secondary families [Efron and Tibshirani, 1994, Holm, 1979, Benjamini and Hochberg, 1995]. Architecture, search, and optimizer settings are in the appendix.

| Game | Training | Eval. | n | Behavioral response | | Policy and search quantities | | |
|---|---|---|---|---|---|---|---|---|
| | | | | Defense | Offense | Net. entropy | Net. tactical mass | Visit tactical mass |
| Gomoku | Unguided | Prior-Off | 3,831 | 0.138 | 0.128 | 2.570 | 0.025 | 0.050 |
| | Unguided | Force-On | 3,831 | 1.000 | 1.000 | 2.570 | 0.025 | 0.665 |
| | Soft-Guided | Prior-Off | 3,831 | 0.265 | 0.211 | 2.626 | 0.033 | 0.104 |
| | Force-Guided | Prior-Off | 3,831 | 0.263 | 0.180 | 2.589 | 0.034 | 0.100 |
| | Force-Guided | Force-On | 3,831 | 1.000 | 1.000 | 2.589 | 0.034 | 0.665 |
| Go | Unguided | Prior-Off | 10,194 | 0.006 | 0.004 | 3.679 | 0.003 | 0.003 |
| | Unguided | Force-On | 10,194 | 0.994 | 0.996 | 3.679 | 0.003 | 0.718 |
| | Soft-Guided | Prior-Off | 10,194 | 0.316 | 0.324 | 3.056 | 0.215 | 0.231 |
| | Force-Guided | Prior-Off | 10,194 | 0.338 | 0.337 | 2.973 | 0.240 | 0.243 |
| | Force-Guided | Force-On | 10,194 | 0.996 | 0.997 | 2.973 | 0.240 | 0.719 |

Table 1: Sealed one-shot final test at checkpoint 40, pooled over three formal seeds; the five rows per game are the only arm × mode combinations the final test contains, and all were queried once. Learned residuals are the Soft-Guided and Force-Guided rows minus the Unguided row within the Prior-Off mode. Holding the network fixed, network entropy and network tactical mass are invariant to the evaluation mode whereas visit tactical mass moves—the signature of an intervention confined to root search. Defense is the confirmatory endpoint; offense and the policy quantities are secondary.

# 4 Results

## 4.1 A Search Prior Leaves a Large but Partial Residual

Table 1 reports the sealed final test. Under Prior-Off—no tactical computation available at inference—Force-Guided raises defensive obligation response from 0.138 to 0.263 in Gomoku (+12.6pp) and from 0.006 to 0.338 in Go (+33.2pp). Soft-Guided yields +12.7pp and +31.1pp. Because every learned comparison fixes evaluation at Prior-Off, none of these gains can come from evaluation-time root restriction; they are properties of the weights.

Two features matter more than the size. First, the curricula agree to within 2.2 points even though they differ sharply as search procedures: Soft-Guided never removes a legal action, yet teaches the defensive obligation about as well as an operator that deletes every alternative. Whatever transports the behavior, it is not the pruning of the action set. Second, the effect is 2.6 times larger in Go than in Gomoku from a far lower base (0.6% versus 13.8%)—which we return to below.

The same table bounds the claim. Applying Force-On to Unguided produces 1.000 in Gomoku and 0.994 in Go; applying it to Force-Guided gives 1.000 and 0.996. The dependence gaps $\Delta_{\mathrm{dep}}$ are therefore 73.7 and 65.8 points: after 40 iterations of curriculum, roughly six-sevenths of the Gomoku behavior and two-thirds of the Go behavior are still supplied online rather than stored. As a share of what the module contributes without touching the weights, $\Delta_{\mathrm{force}}/\Delta_{\mathrm{online}}$ is 0.15 in Gomoku and 0.34 in Go; we quote this ratio descriptively and state all results as differences.

Equation 7 makes the cost of the conventional comparison concrete. The diagonal comparison between Unguided under Prior-Off and Force-Guided under Force-On is +86.2pp in Gomoku, of which only +12.6pp survives removal of the prior, and +99.1pp in Go, of which +33.2pp survives: the diagonal alone overstates learned progress by factors of 6.9 and 3.0. Nor is the interaction negligible—I is −12.6pp and −33.0pp, because the forcing prior has less left to add once the network has absorbed part of the behavior—so additivity would have been the wrong default.

## 4.2 What Changed Inside the Policy

Because Table 1 reports network and search quantities in the same matched cells, it also indicates where the two games store the residual—and the answer differs qualitatively. In Go the transfer is a wholesale reallocation of probability mass. Network tactical mass—the prior mass the unaided network places on oracle-valid tactical actions—rises from 0.003 for Unguided to 0.215 and 0.240 for the guided arms, an increase of +0.212 and +0.237 from a near-zero base, and root visit mass under Prior-Off follows it almost exactly (0.003 to 0.231 and 0.243). The guided Go networks have learned to propose capture and rescue moves; search then merely confirms them.

In Gomoku the same behavioral gain is achieved with a far smaller redistribution: network tactical mass moves only from 0.025 to 0.034, less than one percentage point, while defensive response nearly doubles. The mechanism here is a reordering near the policy's argmax rather than a mass transfer—guidance changes which of several already-plausible moves ranks first, and 32 PUCT simulations amplify that small margin into a different chosen action. This also explains the cross-game asymmetry in effect size: where the baseline already assigns non-trivial mass to the right region (Gomoku, 0.025 with a 13.8% base rate) there is little room for a mass-level effect, whereas where it assigns essentially none (Go, 0.003 with a 0.6% base rate) guidance must build the preference from scratch.

The Go base rate is worth stating plainly rather than defending. Forty iterations of 32 episodes is about 1,300 self-play games on 19×19 with a 4-block network, so the unguided Go arm is a very weak player and 0.6% is what one should expect of it; the same positions are answered 99.4% of the time once the prior is switched on, so they are neither unsolvable nor mislabeled. This is a small-compute regime, not a claim about strong Go agents—but for our purposes it is the cleaner of the two measurements, because a base rate of essentially zero leaves no pre-existing competence to be mistaken for an internalized one.

The policy quantities also close the entropy loophole. If guided training merely sharpened the policy, response would rise with confidence; it does not. In Gomoku the most entropic arm (Soft-Guided, 2.626 nats) and the intermediate one (Force-Guided, 2.589) reach the same response, while Unguided is the least entropic (2.570) and the worst. In Go the ordering reverses: the guided arms are markedly less entropic (2.973, 3.056 versus 3.679) and much more accurate. No monotone entropy–response relation survives both games, which is what a targeted change in action preference—rather than a global change in confidence—should look like.

Finally, the invariance columns provide an implementation check: within a trained arm, switching Prior-Off to Force-On leaves both network quantities unchanged to every reported digit—bitwise, in the archive—while lifting visit tactical mass by 61.5 points in Gomoku and 71.4 in Go for Unguided, but only 56.4 and 47.5 for Force-Guided, which already directs some visits there on its own.

## 4.3 The Residual Emerges with Training and Does Not Transfer

The final test is a single time point. To watch the residual appear—and to test exchangeability behaviorally—Fig. 2 tracks defensive response under Prior-Off across all six checkpoints of the matched two-layer evaluation, where all three arms are scored on identical positions. Being one seed of one game, it corroborates the three-seed final test rather than independently confirming it.

The in-distribution panel shows the pattern that an internalization account predicts and an artifact account does not. At checkpoint 0 the arms are within 1.1 points of one another (4, 7, and 6 correct out of 271), so the shared initialization is behaviorally as well as structurally verified. The arms then separate in the expected order, though not monotonically: by checkpoint 40, Force-

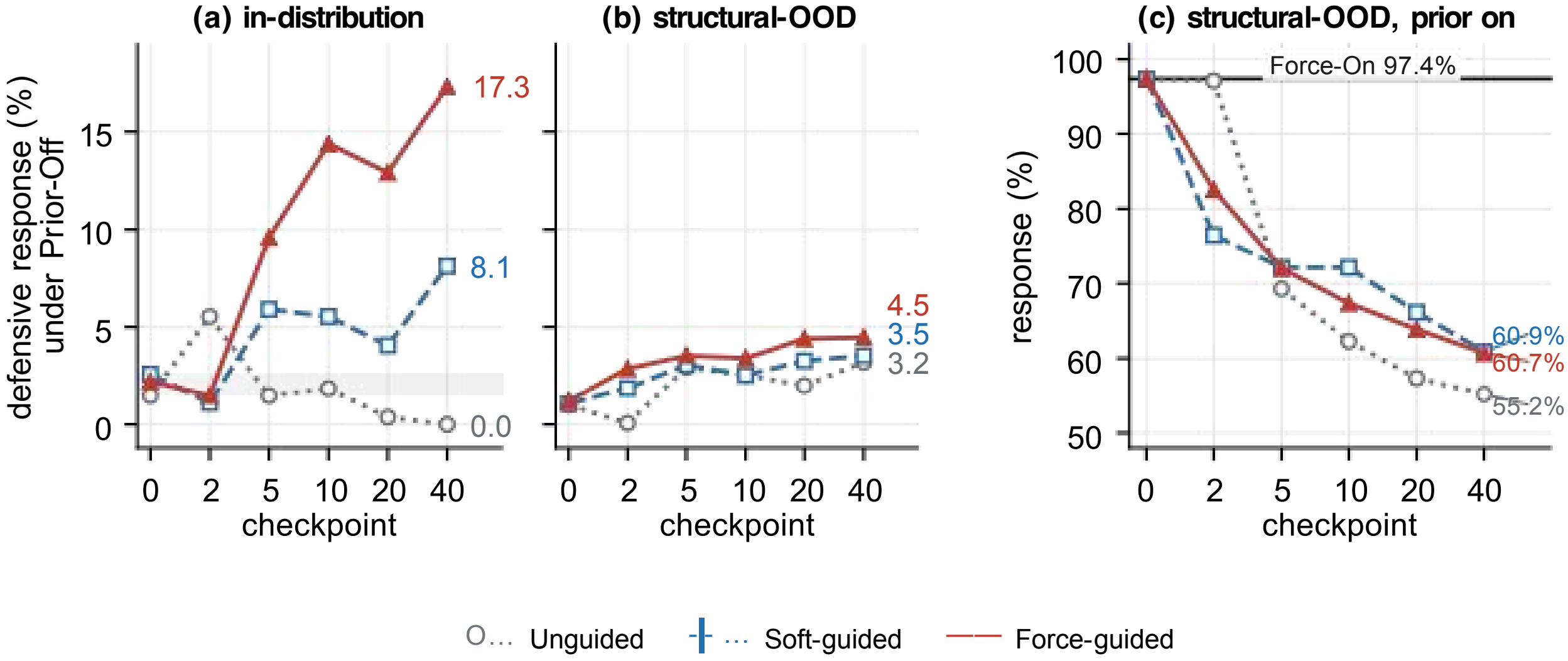

Figure 2: Matched-position defense evaluation (Gomoku, seed 1): all three arms are scored on identical positions at every checkpoint (n = 271 in-distribution, n = 1,500 structural-OOD, per arm per checkpoint). Panels (a) and (b) share a vertical scale and are measured under Prior-Off, so every value plotted there is learned; the band marks the checkpoint-0 spread across arms. (a) The residual is absent at the shared initialization and emerges over training—not monotonically—reaching 17.3% for Force-Guided against 0.0% for Unguided (47/271 versus 0/271); the Unguided rise to 5.5% at checkpoint 2 shows how far one arm moves on a set this small. (b) On structurally novel templates the curricula separate by only 1.3 points (67 versus 48 of 1,500) and all arms stay below 5%. (c) The same OOD positions with a prior on at test time, on its own scale: Force-On holds 1,461/1,500 for every arm and checkpoint, so neither the ceiling nor detectability limits (b), while Soft-On starts at that ceiling and decays to 55–61%—the two operators are far from interchangeable at inference though nearly equivalent as curricula.

Guided reaches 17.3% (47/271) and Soft-Guided 8.1% (22/271), while Unguided falls to 0.0% (0/271). Single-arm movement between adjacent checkpoints is itself several points on a set of 271—Unguided peaks at 5.5% at checkpoint 2—so on this panel only the Force-Guided margin is comfortably above that scale, and the Soft-Guided residual rests on the far larger final-test cells. The baseline's decline is informative in its own right: unguided self-play does not merely fail to acquire the defensive obligation, it drifts away from a behavior its random initialization exhibited by chance, because nothing in unguided self-play makes rare forced defenses worth sampling. Guidance does not accelerate a behavior that would otherwise appear later—it is why the behavior appears at all.

The structural-OOD panel is where the claim stops. On tactical geometries that no training family contained, the curricula separate by only 1.3 points (67 versus 48 out of 1,500), a thirteenfold reduction in margin, and every arm remains below 5%. The ceiling is not responsible: under Force-On all arms answer 1,461 of 1,500 of these same positions correctly, so the obligations are detectable and the correct actions available. What the guided networks acquired is therefore much closer to a set of trained motifs than to a transferable rule. This is the sharpest bound in the paper, and one the diagonal ablation cannot even express: an agent that keeps its forcing prior at test time would score above 97% on exactly these positions and appear to generalize perfectly.

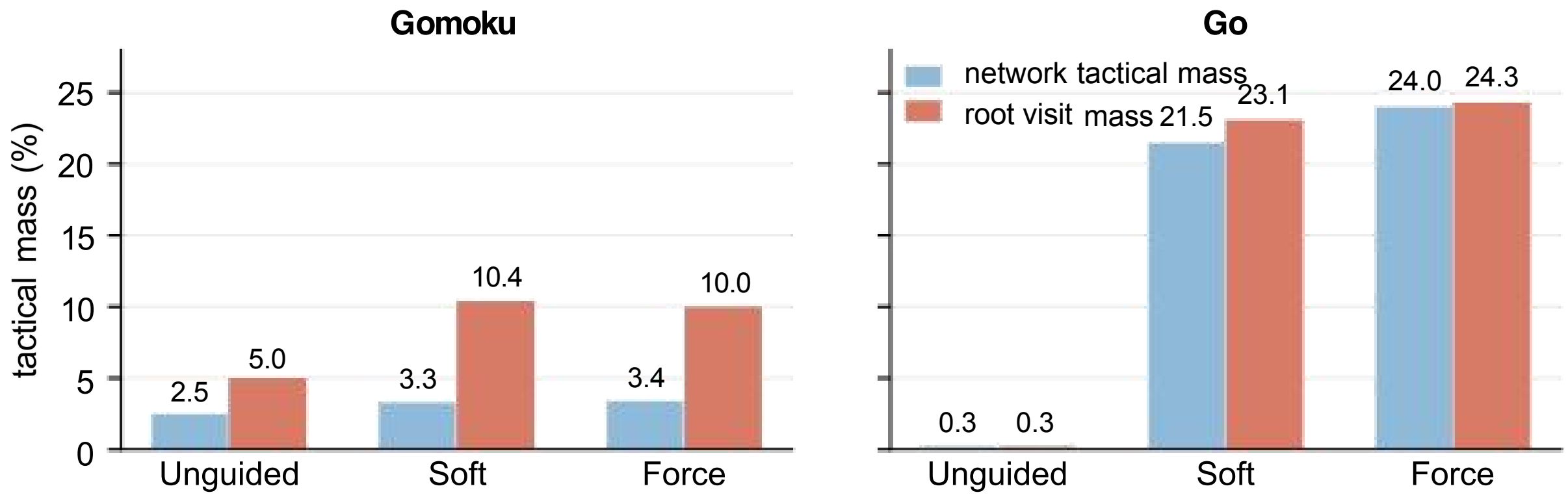

Figure 3: Secondary policy analysis at checkpoint 40. Network tactical mass is the unaided policy probability assigned to oracle-valid tactical actions; root visit mass is the corresponding fraction after 32-search simulation. The quantities are already reported in Table 1; this visualization is descriptive and does not define an additional confirmatory endpoint. The small Gomoku change is consistent with near-argmax reordering, whereas Go shows a broader transfer of probability mass.

## 4.4 Verification Checks and Alternative Explanations

**Implementation checks.** Across all 31,878 Prior-Off defense records of the matched evaluation — three arms × six checkpoints × 1,771 positions—the intervention's call counter is zero, so no evaluation labeled "prior off" silently invoked the module. Under Force-On the attainable ceiling is 100% in-distribution and 97.4% on the OOD layer, the shortfall being positions where the module detects no forcing move.

**Is this just imitation of a deterministic label?** If the network were copying pseudo-labels, the operator that supplies them most aggressively should teach best and the student should approach the teacher. Neither happens: Soft-Guided never deletes a legal action yet matches Force-Guided on the confirmatory endpoint, and Force-Guided reaches only 0.263 and 0.338 against a Force-On ceiling near 1.0. Transfer is indirect, mediated by search visits, and heavily lossy.

**Are the two operators simply the same intervention?** The most economical way to explain Soft-Guided matching Force-Guided would be that reweighting and restriction differ only cosmetically. The matched evaluation says otherwise: switched on at test time on the structural-OOD layer, Force-On holds 1,461/1,500 for every arm at every checkpoint, whereas Soft-On begins at that same ceiling and decays to 55–61% by checkpoint 40 (Fig. 2(c)). A bounded logit bonus loses its grip precisely as a network's own prior sharpens enough to outvote it; a mask cannot lose its grip at all. The two operators are therefore sharply different as inference procedures and nearly interchangeable as curricula—a dissociation that is itself the result, not an artifact of their being one knob.

**Is it memorization?** Family disjointness across splits and the structural-OOD layer make exact template reuse unlikely, but they do not exclude reuse of local motifs, and the collapse of the margin from 17.3 to 1.3 points is precisely what partial motif transfer predicts. We therefore call the residual motif-level rather than rule-level. Excluding motif reuse would require adversarially transformed but legal obligations, or intervention on internal representations.

## 5 Search as Teacher: Two Pathways

CPI identifies a total behavioral residual, and the AlphaZero update suggests two channels through which it can arise. Let $H$ denote a root intervention, $\theta^-$ the behavior parameters that generated the replay buffer, $d_{\theta^-,H}$ the induced state distribution, and $\pi_{\theta^-,H}$ the normalized root-visit target. Ignoring value and regularization terms, one policy update minimizes

$$\mathcal{L}_H(\theta;\theta^-) = -\mathrm{E}_{s\sim d_{\theta^-,H}} \sum_{a\in A(s)} \pi_{\theta^-,H}(a \mid s) \log p_\theta(a \mid s), \tag{8}$$

where $\pi_{\theta^-,H}$ is extended by zero outside the root action set $H$ leaves available, so the sum runs over all legal actions in both conditions. Changing $H$ changes the target inside the expectation and the state distribution outside it: target teaching and trajectory teaching.

Target teaching is local. Reweighting or restricting the root moves visit counts at a fixed state, and the policy head is trained toward the result; repeated exposure produces a residual on states of that kind. Our mass measurements are its fingerprint, in two regimes: broad reallocation in Go (0.003 to 0.240 under Prior-Off) and narrow reordering near the argmax in Gomoku (0.025 to 0.034, with a much larger response change).

Trajectory teaching is global. The guided action also changes the next board, hence every subsequent training position and value target—defensive guidance prolongs games and exposes counterattacks, forcing offense shortens them and removes quiet positions—so two agents accumulate divergent replay buffers even where their targets agree on shared states. This may teach behavior the module never labels, but it may equally narrow coverage, which is the natural reading of Fig. 2(b).

Equation 8 also specifies the experiment that would separate the channels: cross the state source $d \in \{d_0, d_H\}$ with a recomputed root target $\pi \in \{\pi_0, \pi_H\}$ and train the four resulting students, matching state multiplicities and value targets so the construction does not smuggle the shift back in. We estimate the total effect because that is what a practitioner receives, but the split determines the remedy: distill more root targets if target teaching dominates, diversify guided trajectories or mix unguided replay if coverage does.

Even a perfectly optimized Eq. 8 would not reproduce the teacher. The network is fit on a finite on-policy state distribution, whereas Force-On applies a rule to novel states—exactly the gap Fig. 2(b) exposes—and a discontinuous action mask can only be approximated by a continuous policy. The dependence gap is therefore not merely failed learning; it measures the standing advantage of a procedure that generalizes by construction rather than by interpolation.

## 6 Implications for Search-Guided Learning

Whenever a component participates in training-time action selection, its ablation must be crossed rather than diagonal. A diagonal comparison mixes inference correction with learning, and Eq. 7 shows the mixture is not even additive; an inference-only comparison makes the opposite error, holding the network fixed and so missing that guided self-play produced a different network. A complete evaluation reports the cross-phase matrix rather than its corners, verifies the component is provably inert when off, presents learned residuals beside the remaining online dependence, and freezes an independent oracle before the final test opens.

It also follows that "better search" and "better network" are different products. Removing the teacher tests compression into weights; retaining it tests a hybrid in which symbolic and learned components cover each other's failures. Here the hybrid is the stronger engineering choice—the dependence gap is 66 to 74 points—and what is not defensible is using hybrid online performance as evidence that compression succeeded. The soft-versus-hard comparison adds a two-sided corollary. As a curriculum, soft reweighting nearly matched hard restriction at lower risk, since it never deletes a legal action: a flaw in the forcing-set design cannot harden into a constraint, and the exploration a restricted root destroys is preserved. As an inference-time guarantee it is the weaker, surrendering some 37 points of OOD enforcement by checkpoint 40. Teaching a behavior and enforcing it call for different operators.

# 7 Limitations

This study focuses on behavioral internalization, demonstrating that prior-deleted networks exhibit a statistically significant preference for oracle-valid tactical actions. We acknowledge that this behavioral evidence does not address the network's internal representation of heuristic concepts—a question that is better suited for representation-level interventions. Furthermore, while tactical validity is a necessary condition for strong gameplay, it is not sufficient, as a locally sound defense may still be vulnerable to long-term strategic exploitation. Therefore, we position our tactical analysis as a diagnostic foundation, and view full-game evaluation as a complementary, rather than competitive, layer of validation.

# 8 Conclusion

A root prior that alters neither reward, observation, architecture, nor loss still leaves a decisive imprint on the learned weights. Remove it, and defensive competence nearly doubles in Gomoku and surges from near zero in Go—a striking testament to its leverage. Soft reweighting proves nearly as effective as hard pruning, revealing that what truly educates the network is the reallocation of search effort, not the suppression of alternative moves.

Yet this influence remains both partial and parochial. Between two-thirds and six-sevenths of the resulting behavior is still borrowed from search, and the hard-won advantage all but evaporates when confronted with unseen board geometries. Search in self-play, then, is not merely planning by another name. Root priors reshape visits; visits reshape targets; targets reshape weights; and weights, in turn, reshape the very states the agent encounters. Policy improvement, therefore, should ultimately be measured not by what the student performs under the teacher's guidance, but by what it retains once the teacher has left the room.

# A Formal Specification of the Guidance Operators

## A.1 Tactical Module

Let $A(s)$ be the legal actions at root state $s$ and $p_\theta(\cdot \mid s)$ the network prior renormalized on $A(s)$. The tactical module is a deterministic function of the board alone. It returns a threat score $r(s, a)$ and a forcing set $M(s) \subseteq A(s)$.

For a candidate action $a$, let $\mathrm{atk}(s, a)$ be the value of the strongest pattern $a$ creates for the mover and $\mathrm{dfn}(s, a)$ the value of the strongest pattern $a$ denies the opponent. Both are read from the same tier table, so blocking a threat is scored at the value of the threat blocked:

$$r(s, a) = \max\{\mathrm{atk}(s, a),\ \mathrm{dfn}(s, a)\} \in [0, W]. \tag{9}$$

Table 2 lists the tier tables and constants for both games. Compound-threat detection and tier ablation hooks exist in the implementation but are disabled in every run reported in the paper, so $r$ takes values in the finite tier set of Table 2.

The forcing sets are game-specific, and this is deliberate. Let $C_{\mathrm{cap}}(s) \subseteq A(s)$ be the points that capture an opponent group in atari and $C_{\mathrm{save}}(s) \subseteq A(s)$ the points that raise a mover group out of atari. Then

$$M_{\mathrm{Gomoku}}(s) = \{a \in A(s) : r(s, a) \geq \tau\}, \tag{10}$$

$$M_{\mathrm{Go}}(s) = \bigcup_{\substack{t \in \{\mathrm{cap}, \mathrm{save}\} \\ |C_t(s)| = 1}} C_t(s). \tag{11}$$

In Gomoku every action at or above the tier $\tau$ is forcing, so $M(s)$ may contain several equally valid blocks. In Go a tier enters $M(s)$ only when it has a unique candidate. The reason is that hard restriction deletes actions: if two distinct stones could be captured, neither capture is obligatory, and pruning to one of them would fabricate an obligation the rules do not impose. This asymmetry is a property of the intervention, not of the oracle: the outcome labeler always returns the complete set of correct actions (Section D). Note that $M(s) \subseteq \{a : r(s, a) > 0\}$ in both games, since every tier in Table 2 is strictly positive; Eq. 16 below relies on this.

Given $r$ and $M$, the logit bonus is

$$b(s, a) = \frac{\lambda}{W}\Big[r(s, a) + \tau\, \mathbf{1}\{a \in M(s)\}\Big]. \tag{12}$$

The constant $\tau$ plays two distinct roles, and it is worth separating them. In Eq. 12 it is the size of the bonus awarded for membership in $M(s)$, in both games. In Gomoku it additionally serves as the threshold that defines $M(s)$ in Eq. 10; in Go it does not, because Eq. 11 is defined by uniqueness, so a Go rescue with $r = 200 < \tau$ can still belong to $M(s)$. Two further consequences make the operators below well defined. First, $b(s, a) = 0$ whenever $a$ has no tactical content, so a tactically empty position leaves the network prior exactly unchanged. Second, $b$ is bounded: $0 \leq b(s, a) \leq \lambda(1+\tau/W)$, which is 2.4 in Gomoku and 4.0 in Go.

## A.2 The Three Root Operators

Each evaluation mode is a map from the network prior to a root prior $q_g(\cdot \mid s;\theta)$ supported on a root action set $A_g(s)$. Prior-Off is the identity,

$$q_O(a \mid s;\theta) = p_\theta(a \mid s), \qquad A_O(s) = A(s). \tag{13}$$

| Symbol | Meaning | Gomoku | Go |
|---|---|---|---|
| λ | bonus strength | 2.0 | 2.0 |
| W | value of an immediate win | 1000 | 1000 |
| τ | membership bonus | 200 | 1000 |
| Tier table for atk and dfn | | | |
| | five in a row / group capture | 1000 | 1000 |
| | open four / atari rescue | 200 | 200 |
| | simple four | 80 | — |
| | open three | 40 | — |

Table 2: Constants of the tactical module. Numerically τ equals the open-four tier in Gomoku and the capture tier in Go, but it acts only as the membership bonus of Eq. 12 in both games and as the threshold of Eq. 10 in Gomoku alone. Identical values are used during training and during evaluation, so the training and evaluation switches differ only in when the operator is applied.

Soft-On tilts the prior while preserving support,

$$q_S(a \mid s;\theta) = \frac{p_\theta(a \mid s)\, e^{b(s,a)}}{\sum_{u \in A(s)} p_\theta(u \mid s)\, e^{b(s,u)}}, \quad A_S(s) = A(s). \tag{14}$$

For Force-On, write $R_M(s) = \sum_{u \in M(s)} r(s,u)$ and define the forcing-set distribution

$$\mu_M(a \mid s) = \begin{cases} r(s,a)/R_M(s), & a \in M(s),\ R_M(s) > \epsilon, \\ 1/|M(s)|, & a \in M(s),\ R_M(s) \le \epsilon, \\ 0, & a \notin M(s), \end{cases} \tag{15}$$

with $\epsilon = 10^{-8}$. Then

$$q_F(a \mid s;\theta) = \begin{cases} \mu_M(a \mid s), & M(s) \neq \emptyset, \\ q_S(a \mid s;\theta), & M(s) = \emptyset, \end{cases} \tag{16}$$

with $A_F(s) = M(s)$ when $M(s) \neq \varnothing$ and $A_F(s) = A(s)$ otherwise.

**Well–posedness.** Each of Eqs. 13–16 is a probability distribution on its stated support. In Eq. 14 the denominator is strictly positive because $p_\theta$ is normalized on A(s) and $e^b > 0$; the implementation clips network probabilities to $[10^{-8}, 1]$ before taking logarithms, which changes no support. In Eq. 15 the two nonzero branches are exhaustive on a nonempty M(s) and each normalizes to one. The second branch is in fact unreachable under the tier tables of Table 2: every $a \in M(s)$ has $r(s, a) \geq 200$, so $R_M(s) \geq 200 \gg \epsilon$ whenever $M(s) \neq \varnothing$. We state it only because the implementation carries the guard, and the main paper's Eq. 3 correspondingly omits it. When $M(s) = \varnothing$, Eq. 16 coincides with Eq. 14 and the indicator in Eq. 12 is inactive, so $b(s, a) = \lambda r(s, a)/W$ there.

**Two properties used in the main paper.** (i) On a nonempty forcing set, $q_F$ does not reweight $p_\theta$; it replaces it by $\mu_M$ and narrows the root action set to M(s). Because $\mu_M$ is supported only on oracle-accepted actions (Section D), near-perfect Force-On response is guaranteed by construction and is reported only as a positive control and as the reference level for the dependence gap. (ii) Guidance is confined to the root. Internal tree nodes always expand with $p_\theta$, and r, b, M never enter the input planes, the value head, or the loss. Consequently, for a fixed θ, switching g leaves all network-side quantities bitwise unchanged and alters only visit counts and the selected action—an invariance we verify numerically in the main paper.

### A.3 Search Targets and Selected Actions

With simulation budget $B$, let $N^B_g(s, a;\theta)$ be the root visit count produced by PUCT from $q_g$. The normalized search target and the greedy action are

$$\pi^B_{\theta,g}(a \mid s) = \frac{N^B_g(s,a;\theta)}{\sum_{u\in A_g(s)} N^B_g(s,u;\theta)}, \tag{17}$$

$$\hat{a}_{\theta,g}(s) = \arg\max_{a\in A_g(s)} N^B_g(s, a;\theta). \tag{18}$$

During self-play, Dirichlet exploration noise is mixed into the root prior after the operator is applied, on the possibly restricted set $A_g(s)$; evaluation is noise-free and uses Eq. 18 with the budget of Table 3.

### A.4 Training Objective and the Two Teaching Channels

Fix a training curriculum $c \in \{U, S, F\}$ and let $H_c \in \{O, S, F\}$ be the operator it applies during self-play. A behavior network $\theta^-$ generates an on-policy distribution $d_{\theta^-,H_c}$ over pairs $(s, z)$ of visited states and terminal outcomes, together with the fixed targets $\pi^B_{\theta^-,H_c}$. One optimization phase minimizes

$$\begin{aligned} \mathcal{L}_{H_c}(\theta;\theta^-) = \mathrm{E}_{(s,z)\sim d_{\theta^-,H_c}} \Big[ -\sum_{a\in A(s)} \pi^B_{\theta^-,H_c}(a \mid s)\log p_\theta(a \mid s) \\ + \big(v_\theta(s) - z\big)^2 \Big] + \beta\|\theta\|_2^2, \end{aligned} \tag{19}$$

where $\pi^B_{\theta^-,H_c}$ is extended by zero on $A(s) \setminus A_{H_c}(s)$, so the inner sum ranges over all legal actions regardless of whether the operator restricted the root. Equation 19 makes the two channels explicit and is the appendix form of the loss discussed in the main paper. Replacing $H_U = O$ by $H_S$ or $H_F$ changes the target inside the expectation (target teaching) and the state distribution outside it (trajectory teaching). Nothing else in Eq. 19 depends on the intervention: the loss form, the value target $z$, and the weight decay $\beta$ are identical across arms, and the module's outputs are never an argument of $p_\theta$ or $v_\theta$. This is the precise sense in which any surviving behavioral difference must have been transported through $\pi^B$ and $d$.

## B Estimands and the Decomposition Identity

Let $\mathcal{D}$ be the pooled index of applicable seed–position pairs $(j, s)$, let $Y(s) \subseteq A(s)$ be the complete set of correct actions returned by the independent oracle, and let $\theta_{c,j,k}$ be the checkpoint-$k$ network of curriculum $c$ under seed $j$. The reported response is

$$R^g_c(k) = \frac{1}{|\mathcal{D}|} \sum_{(j,s)\in\mathcal{D}} \mathbf{1}\big\{\hat{a}_{\theta_{c,j,k},g}(s) \in Y(s)\big\}. \tag{20}$$

Scoring membership in $Y(s)$ rather than equality with a single key is necessary: several defenses are frequently valid at once. Abbreviating $R^g_c \equiv R^g_c(40)$, the pre-declared contrasts are

$$\begin{aligned} \Delta_{\text{online}} &= R^F_U - R^O_U, & \Delta_{\text{soft}} &= R^O_S - R^O_U, \\ \Delta_{\text{force}} &= R^O_F - R^O_U, & \Delta_{\text{dep}} &= R^F_F - R^O_F. \end{aligned} \tag{21}$$

Both learned residuals hold $g = O$ fixed, where the module is provably inert by Eq. 13 and verified to issue zero calls, so neither can absorb an online contribution.

| Setting | Gomoku | Go |
|---|---|---|
| Board | 9×9 | 19×19 |
| Input planes | 17 | 17 |
| Network | 8 blocks, 512 ch. | 4 blocks, 64 ch. |
| Training search B | 64 simulations | 128 simulations |
| Evaluation search B | 32 simulations | 32 simulations |
| Episodes/iteration | 80 | 32 |
| Minibatch size | 2,048 | 1,024 |
| Epochs/iteration | 6 | 6 |
| Training iterations | 40 | 40 |
| Formal seeds | 3 | 3 |
| Training arms per seed | 3 | 3 |

Table 3: Matched-compute configuration within each game. Evaluation compute is identical across arms and modes, so no contrast in Eq. 21 is confounded with search budget.

**The identity.** All four contrasts are differences among the same four cells $R^{O}_{U}$, $R^{F}_{U}$, $R^{O}_{F}$, $R^{F}_{F}$. Adding and subtracting $R^{O}_{F}$ gives the first line below; adding and subtracting $\Delta_{\text{online}}$ gives the second:

$$R^{F}_{F} - R^{O}_{U} = \underbrace{\left(R^{O}_{F} - R^{O}_{U}\right)}_{\Delta_{\text{force}}} + \underbrace{\left(R^{F}_{F} - R^{O}_{F}\right)}_{\Delta_{\text{dep}}}$$

$$= \Delta_{\text{force}} + \Delta_{\text{online}} + I, \tag{22}$$

with $I := \Delta_{\text{dep}} - \Delta_{\text{online}}$. The derivation is exact and assumption-free—it is bookkeeping on four numbers—which is why the interaction I cannot be dropped by appealing to approximate additivity. Substantively, I is the difference between the online value of the forcing prior on a guided network and on an unguided one. It is generically nonzero, and its sign is informative: $I < 0$ means the prior has less left to contribute once part of the behavior is already in the weights, which is what we observe in both games. A corollary used in the main paper: one cannot recover $\Delta_{\text{force}}$ from a published diagonal number by subtracting a separately measured baseline inference gain, because that procedure silently sets $I = 0$.

# C Experimental Configuration

Table 3 gives the complete configuration. Compute is matched across Unguided, Soft-Guided, and Force-Guided within a game and seed; the three arms' configuration objects are verified to be identical except for the two intervention flags. Gomoku and Go architectures differ by design and no causal contrast is ever formed across games. Raw checkpoint indices $\{0, 2, 5, 10, 20, 39\}$ are displayed as completed checkpoints $\{0, 2, 5, 10, 20, 40\}$.

## C.1 Compute Environment

All confirmatory training and evaluation ran on a single NVIDIA Tesla V100-SXM2 (32 GB) under Linux, with Python 3.10 and PyTorch with CUDA acceleration; self-play, optimization, and evaluation all execute on that one device, and the tactical module runs on CPU. Because every contrast is formed within a game and seed on the same device, arm-to-arm timing variation cannot enter any reported effect. Indicative costs on this device: roughly 6 GPU-hours per Gomoku training arm, and roughly 170 seconds per evaluation cell of the defense suites, so the seed-1 matched evaluation of 216 cells took about 5 GPU-hours and the 108-cell cross-evaluation about 66 minutes. The 15 confirmatory training runs and their evaluations together cost on the order of $10^2$ GPU-hours; this single-device budget, not any methodological choice, is why the design uses three formal seeds rather than the ten or more that a variance estimate would need.

| Archive component | Gomoku | Go | Total |
|---|---|---|---|
| Decision records | 539,661 | 161,822 | 701,483 |
| Correct labels | — | | 317,697 |
| Incorrect labels | — | | 216,377 |
| Non-applicable | — | | 167,409 |
| Final-test records | 19,155 | 50,970 | 70,125 |
| arm × mode combinations | 5 | 5 | 10 |
| seed-level cells (5 × 3 seeds) | 15 | 15 | 30 |

Table 4: Evaluation archive sizes. The sealed final test contains exactly the five pre-registered training-arm × evaluation-mode combinations of Section E, recorded once per formal seed, and the single query opened all five; we make no claim to have left further final-test cells unopened. The remaining cells of the 3 × 3 cross—in particular the whole Soft-On row—exist only in the development and validation archives, which were read freely before the analysis lock. Label totals are reported globally because applicability differs by tactical family.

# D Datasets, Labels, and Label Independence

Gomoku obligations are win, block, make-open-four, and block-open-four; Go obligations are capture and save. Position families are canonicalized under the game's symmetry group before splitting, and no family crosses the development, validation, and test partitions.

Defensive validation is two-layered and the layers are never pooled. The in-distribution layer draws held-out instances of trained families. The structural-OOD layer contains geometries that appear in no training or development family: sealed consecutive fours, broken fours, corner obligations, multi-threat states, multi-stone atari chains, counter-captures, and positions admitting several valid saves.

The outcome labeler lives in a module that is statically forbidden from importing the intervention code. It receives a board, a legal-action set, and a requested obligation, and returns the complete set $Y(s)$ of correct actions derived from the game rules. Golden tests verify one direction only: every action the intervention would force must be legal and must belong to $Y(s)$. The converse is deliberately not required, so $Y(s)$ may be strictly larger than $M(s)$—for instance when several blocks are equally valid, or when a Go tier is non-unique and therefore excluded from $M(s)$ by Eq. 11. Without this one-sided relation, near-perfect Force-On response would be circular.

# E Pre-Registered Contrasts and Guardrails

The five frozen primary cells per game are Unguided/Prior-Off, Unguided/Force-On, Soft-Guided/Prior-Off, Force-Guided/Prior-Off, and Force-Guided/Force-On. These are exactly what Eq. 21 requires, and the final test was built to contain only them (Table 4). Soft-On cells were used during development and validation and were excluded from the confirmatory protocol.

Defensive obligation response is the single confirmatory endpoint. Offensive families, policy entropy, and tactical-mass quantities are secondary and reported descriptively. In-distribution defense, structural-OOD defense, and the final test are reported separately and never averaged into one generalization number.

The frozen analysis plan specifies hierarchical resampling—seeds first, then position or template families within seed— with Holm correction across the primary contrasts and Benjamini– Hochberg control across secondary endpoints. As stated in the main paper, three formal seeds do not support a credible variance estimate for a clustered binary outcome, so the confirmatory results are reported as point estimates with exact cell sizes and are calibrated against the checkpoint-0 cross-arm spread, which is at most 1.1 points on the matched defense sets. Single-seed pilot runs are retained in the engineering record and excluded from every confirmatory claim.

# F Reproducibility and Verification Protocol

Within a seed, all three arms start from one identical untrained model state, verified by a canonical hash over ordered parameter names, shapes, dtypes, and raw bytes; optimizer state, replay buffer, and random streams restart from the same seed, and no arm is ever warm-started from another. Under Prior-Off the module's call counter is required to be exactly zero, verified over all 31,878 Prior-Off defense records of the matched evaluation (three arms × six checkpoints × 1,771 positions). For a fixed checkpoint, switching Prior-Off to Force-On must leave network policy entropy and network tactical mass bitwise unchanged.

# References

Rishabh Agarwal, Max Schwarzer, Pablo Samuel Castro, Aaron Courville, and Marc G. Bellemare. Deep reinforcement learning at the edge of the statistical precipice. In Advances in Neural Information Processing Systems, volume 34, pages 29304–29320, 2021.

Thomas Anthony, Zheng Tian, and David Barber. Thinking fast and slow with deep learning and tree search. In Advances in Neural Information Processing Systems, volume 30, 2017.

Yoshua Bengio, Jérôme Louradour, Ronan Collobert, and Jason Weston. Curriculum learning. In International Conference on Machine Learning, pages 41–48, 2009.

Yoav Benjamini and Yosef Hochberg. Controlling the false discovery rate: A practical and powerful approach to multiple testing. Journal of the Royal Statistical Society: Series B, 57(1):289–300, 1995.

Cameron B. Browne, Edward Powley, Daniel Whitehouse, Simon M. Lucas, Peter I. Cowling, Philipp Rohlfshagen, Stephen Tavener, Diego Perez, Spyridon Samothrakis, and Simon Colton. A survey of monte carlo tree search methods. IEEE Transactions on Computational Intelligence and AI in Games, 4(1):1–43, 2012.

Karl Cobbe, Oleg Klimov, Chris Hesse, Taehoon Kim, and John Schulman. Quantifying generalization in reinforcement learning. In International Conference on Machine Learning, pages 1282–1289, 2019.

Rémi Coulom. Efficient selectivity and backup operators in monte-carlo tree search. In Computers and Games, pages 72–83, 2006.

Ivo Danihelka, Arthur Guez, Julian Schrittwieser, and David Silver. Policy improvement by planning with Gumbel. In International Conference on Learning Representations, 2022.

Bradley Efron and Robert J. Tibshirani. An Introduction to the Bootstrap. Chapman and Hall/CRC, 1994.

Jean-Bastien Grill, Florent Altché, Yunhao Tang, Thomas Hubert, Michal Valko, Ioannis Antonoglou, and Rémi Munos. Monte-carlo tree search as regularized policy optimization. In International Conference on Machine Learning, pages 3769—3778, 2020.

Peter Henderson, Riashat Islam, Philip Bachman, Joelle Pineau, Doina Precup, and David Meger. Deep reinforcement learning that matters. In AAAI Conference on Artificial Intelligence, volume 32, 2018.

Geoffrey Hinton, Oriol Vinyals, and Jeff Dean. Distilling the knowledge in a neural network. arXiv:1503.02531, 2015.

Sture Holm. A simple sequentially rejective multiple test procedure. Scandinavian Journal of Statistics, 6(2):65—70, 1979.

Thomas Hubert, Julian Schrittwieser, Ioannis Antonoglou, Mohammadamin Barekatain, Simon Schmitt, and David Silver. Learning and planning in complex action spaces. In International Conference on Machine Learning, pages 4476—4486, 2021.

Guido W. Imbens and Donald B. Rubin. Causal Inference for Statistics, Social, and Biomedical Sciences. Cambridge University Press, 2015.

Robert Kirk, Amy Zhang, Edward Grefenstette, and Tim Rocktäschel. A survey of generalisation in deep reinforcement learning. Artificial Intelligence, 320:103923, 2023.

Levente Kocsis and Csaba Szepesvári. Bandit based monte-carlo planning. In European Conference on Machine Learning, pages 282—293, 2006.

M. Pawan Kumar, Benjamin Packer, and Daphne Koller. Self-paced learning for latent variable models. In Advances in Neural Information Processing Systems, volume 23, 2010.

Andrew Y. Ng, Daishi Harada, and Stuart Russell. Policy invariance under reward transformations: Theory and application to reward shaping. In International Conference on Machine Learning, pages 278–287, 1999.

Andrew Patterson, Samuel Neumann, Martha White, and Adam White. Empirical design in reinforcement learning. Journal of Machine Learning Research, 24(359):1—63, 2023.

Judea Pearl. Causality: Models, Reasoning, and Inference. Cambridge University Press, 2 edition, 2009.

Stéphane Ross, Geoffrey J. Gordon, and J. Andrew Bagnell. A reduction of imitation learning and structured prediction to no-regret online learning. In International Conference on Artificial Intelligence and Statistics, pages 627—635, 2011.

Andrei A. Rusu, Sergio Gómez Colmenarejo, Çağlar Gülçehre, Guillaume Desjardins, James Kirkpatrick, Razvan Pascanu, Volodymyr Mnih, Koray Kavukcuoglu, and Raia Hadsell. Policy distillation. In International Conference on Learning Representations, 2016.

Julian Schrittwieser, Ioannis Antonoglou, Thomas Hubert, Karen Simonyan, Laurent Sifre, Simon Schmitt, Arthur Guez, Edward Lockhart, Demis Hassabis, Thore Graepel, et al. Mastering atari, go, chess and shogi by planning with a learned model. Nature, 588:604—609, 2020.

David Silver, Aja Huang, Chris J. Maddison, Arthur Guez, Laurent Sifre, George van den Driessche, Julian Schrittwieser, Ioannis Antonoglou, Veda Panneershelvam, Marc Lanctot, et al. Mastering the game of go with deep neural networks and tree search. Nature, 529(7587):484—489, 2016.

David Silver, Julian Schrittwieser, Karen Simonyan, Ioannis Antonoglou, Aja Huang, Arthur Guez, Thomas Hubert, Lucas Baker, Matthew Lai, Adrian Bolton, et al. Mastering the game of go without human knowledge. Nature, 550(7676):354—359, 2017.

David Silver, Thomas Hubert, Julian Schrittwieser, Ioannis Antonoglou, Matthew Lai, Arthur Guez, Marc Lanctot, Laurent Sifre, Dharshan Kumaran, Thore Graepel, et al. A general reinforcement learning algorithm that masters chess, shogi, and go through self-play. Science, 362(6419): 1140–1144, 2018.

David J. Wu. Accelerating self-play learning in go. arXiv:1902.10565, 2019.

Weirui Ye, Shaohuai Liu, Thanard Kurutach, Pieter Abbeel, and Yang Gao. Mastering atari games with limited data. In Advances in Neural Information Processing Systems, volume 34, pages 25476–25488, 2021.

Chiyuan Zhang, Oriol Vinyals, Rémi Munos, and Samy Bengio. A study on overfitting in deep reinforcement learning. arXiv:1804.06893, 2018.